\pdfoutput=1

\documentclass[11pt]{article}

\usepackage[final]{acl}

\usepackage{times}
\usepackage{latexsym}

\usepackage[T1]{fontenc}

\usepackage[utf8]{inputenc}

\usepackage{microtype}

\usepackage{url}
\usepackage{booktabs}
\usepackage{multirow}
\usepackage{amsmath}
\usepackage{graphicx}
\usepackage{color,soul}
\usepackage{xcolor}

\makeatletter
\newcommand*\bigcdot{\mathpalette\bigcdot@{.7}}
\newcommand*\bigcdot@[2]{\mathbin{\vcenter{\hbox{\scalebox{#2}{$\m@th#1\bullet$}}}}}
\makeatother

\newcommand{\dataset}[1]{{ConditionalQA}} 

\title{ConditionalQA: A Complex Reading Comprehension Dataset \\ with Conditional Answers}

\author{Haitian Sun \\
  School of Computer Science \\
  Carnegie Mellon University \\
  \texttt{haitians@cs.cmu.edu} \\\And
  William W. Cohen \\
  Google Research \\
  \texttt{wcohen@google.com} \\ \And
  Ruslan Salakhutdinov \\
  School of Computer Science \\
  Carnegie Mellon University \\
  \texttt{rsalakhu@cs.cmu.edu} \\
  }

\begin{document}
\maketitle

\begin{abstract}
We describe a Question Answering (QA) dataset that contains complex questions with \textit{conditional answers}, i.e. the answers are only applicable when certain conditions apply. We call this dataset \dataset{}. In addition to conditional answers, the dataset also features:
(1) long context documents with information that is related in logically complex ways;
(2) multi-hop questions that require compositional logical reasoning;
(3) a combination of extractive questions, yes/no questions, questions with multiple answers, and not-answerable questions;
(4) questions asked without knowing the answers.
We show that \dataset{} is challenging for many of the existing QA models, especially in selecting answer conditions. We believe that this dataset will motivate further research in answering complex questions over long documents. Data and leaderboard are publicly available\footnote{\url{https://github.com/haitian-sun/ConditionalQA}}.

\end{abstract}

\section{Introduction}
\begin{figure}
    \centering
    \includegraphics[width=0.48\textwidth]{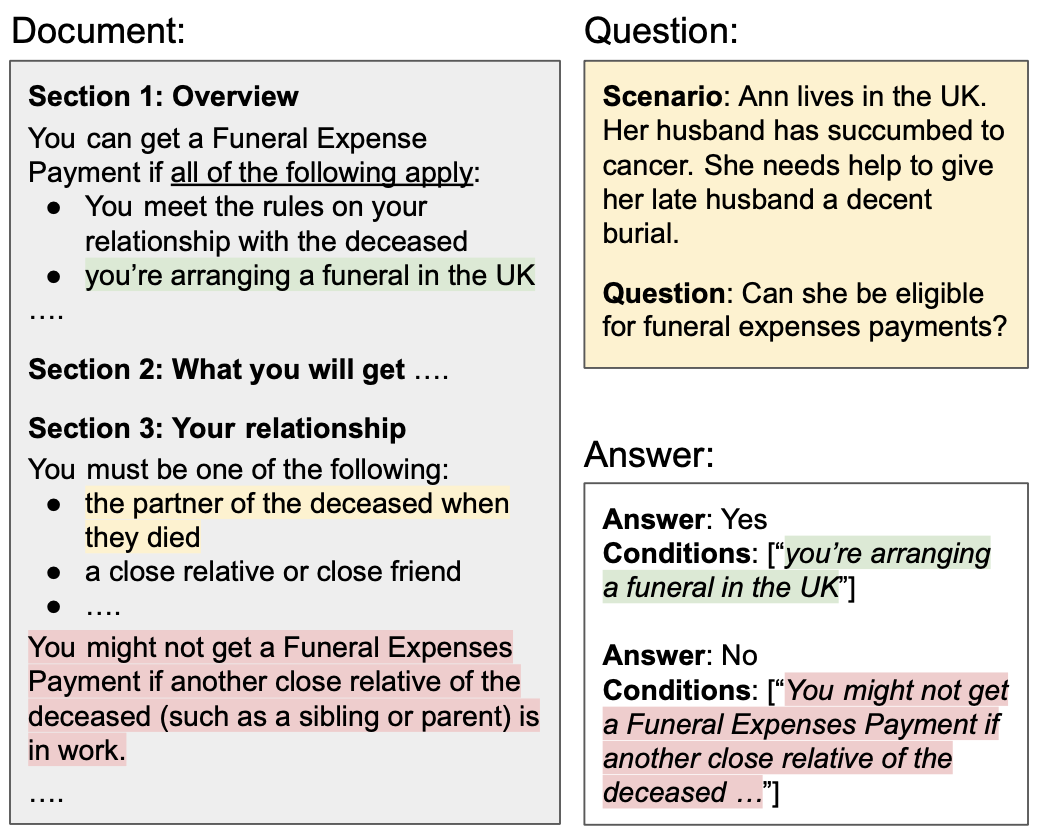}
    \caption{\small An example of question and document in \dataset{} dataset. The left side is a snapshot of the document discussing the eligibility of the benefit ``Funeral Expense Payment''. The text span ``Her husband'' satisfies the requirement on the ``relationship with the decease'' (in yellow).
    Text pieces in green and red are requirements that must be satisfied and thus selected as conditions for the ``Yes'' and ``No'' answers.}
    \label{fig:example}
\end{figure}

Many reading comprehension (RC) datasets have been recently proposed \cite{rajpurkar2016squad, rajpurkar2018know, nq, yang2018hotpotqa, dasigi2021dataset, ferguson2020iirc}. In a reading comprehension task, models are provided with a document and a question and asked to find the answers. Questions in existing reading comprehension datasets generally have a unique answer or a list of answers that are equally correct, e.g. ``Who was the president of the US?'' with the answers ``George Washington'', ``Thomas Jefferson'', etc. We say that these questions have \textit{deterministic} answers. However, questions in the real world do not always have deterministic answers, i.e. answers to the questions are different under different conditions. For example, in Figure \ref{fig:example}, the document discusses ``Funeral Expense Payment'' and a question asks an applicant's eligibility. 
This question cannot be deterministically answered: the answer is ``yes'' only if ``you're arranging a funeral in the UK'', while the answer is ``no'', if ``... another close relative of the deceased is in work'' is true. We call answers that are different under different conditions \textit{conditional answers}.





A conditional answer consists of an \textit{answer} and a list of \textit{conditions}. An answer is only true if its conditions apply. 
In the example above, ``you are arranging a funeral in the UK'' is the condition for the answer ``yes''. An answer can have multiple conditions. Conditional answers are commonly seen when the context is complex; for example, when a person asks a question with some prior knowledge in mind but cannot enumerate all necessary details to get an deterministic answer. We say that such questions are \textit{incomplete}. 
A practical way to answer incomplete questions is to find all possible answers to the question -- and if some answers are only true under certain conditions, the conditions should be output as well.



We present the \dataset{} dataset, which contains questions with conditional answers. Questions are asked about public policies in the UK by local residents. Each example contains a user's question, a description of the scenario that the question is asked, and a long document discussing the policy that the question asks about. The model is asked to read the document and find all possible answers that apply to the user's scenario. If an answer is only true under certain conditions, the model should return the list of conditions along with the answer. 
We provide supporting evidences labeled by human annotators as additional supervision.

In addition to having conditional answers, \dataset{} also features the following properties. 
First, the documents in \dataset{} have complex structure. As opposed to Wikipedia pages, where most sentences or paragraphs contain stand-alone information, documents in \dataset{} usually have complex internal logic that is crucial for answering the questions. 
Second, many questions in the dataset are naturally multi-hop, as illustrated in the example on Figure \ref{fig:example}, e.g. being ``the partner of the deceased'' satisfied the requirement on ``your relationship with the deceased'' which is one of high-level requirements to obtain the benefit. Answering those question requires models that understand the internal logic within the document and reason over the it to find correct answers.
Third, we decouple the asking and answering process when annotating questions, as suggested by \citet{ferguson2020iirc, dasigi2021dataset, clark2020tydi}, so questions are asked without knowing the answers.
Forth, \dataset{} contains various types of questions including yes/no questions and extractive questions. Questions can have one or multiple answers, or can be not answerable, as a result of the decoupled annotation process. 

We experimented with several strong baseline models on \dataset{} \cite{ainslie2020etc, sun2021endtoend, izacard2021leveraging}. The best performing model achieves only 64.9\% accuracy on yes/no questions, marginally better than the majority baseline (62.2\% if always predicting ``yes''), and 25.2\% exact match (EM) on extractive answers. We further measure the accuracy of jointly predicting answers and conditions, in which case the accuracy drops to 49.1\% and 22.5\%. The best results with conditions are obtained if \textit{no condition} is ever predicted, showing how challenging it is for existing models to predict conditional answers. 



\section{Related Works}

Many question answering datasets have been proposed in the past few years \cite{rajpurkar2016squad, rajpurkar2018know, yang2018hotpotqa, dasigi2021dataset, ferguson2020iirc, nq} and research on these has significantly boosted the performance of QA models. As large pretrained language models \cite{devlin2019bert, liu2019roberta, ainslie2020etc, beltagy2020longformer, guu2020realm, verga2020facts} achieved better performance on traditional reading comprehension and question answering tasks, efforts have been made to make the questions more complex. Several multi-hop QA datasets were released \cite{yang2018hotpotqa, ferguson2020iirc, talmor2018web, welbl2018constructing} to test models' ability to solve complex questions. However, most questions in these datasets are answerable by focusing on a small piece of evidence at a time, e.g. a sentence or a short passage, leaving reasoning through long and complex contents a challenging but unsolved problem.


Some datasets have been recently proposed for question answering over long documents. QASPER \cite{dasigi2021dataset} contains questions asked from academic papers, e.g. ``What are the datasets experimented in this paper?''. To answer those questions, the model should read several sections and collect relevant information. NarrativeQA \cite{mou2021narrative} requires reading entire books or movie scripts to answer questions about their characters or plots. Other datasets, e.g. HybridQA \cite{chen2021hybridqa}, can also be viewed as question answering over long documents if tables with hyper-linked text from the cells are flattened into a hierarchical document. ShARC \cite{saeidi2018interpretation} is a conversational QA dataset that requires to understand complex contents to answer questions. However, the ShARC dataset only contains yes/no questions and the conversation history is generated by annotators with the original rule text in hand, making the conversation artificial. The length of context in ShARC is usually short, such as a few sentences or a short paragraph.

Most of the existing datasets, including the ones discussed above, contain questions with unique answers. Answers are unique because questions are well specified, e.g. ``Who is the president of the US in 2010?''. However, questions can be ambiguous if not all information is provided in the question, e.g. ``When was the Harry Potter movie released?'' does not specify which Harry Potter movie. AmbigQA \cite{min2020ambigqa} contains questions that are ambiguous, and requires the model to find all possible answers of an ambiguous question and rewrite the question to make it well specified. Similar datasets Temp-LAMA \cite{dhingra2021timeaware}, TimeQA \cite{chen2021dataset} and SituatedQA \cite{zhang2021situatedqa} have been proposed that include questions that require resolving temporal or geographic ambiguity in the context to find the answers.
They are similar to \dataset{} in that questions are incomplete, but \dataset{} focuses on understanding documents with complex logic and answering questions with conditions. It's usually not possible to disambiguate questions in \dataset{} as rewriting the questions (or scenarios) to reflect all conditions of answers to make the questions deterministic is impractical.


We create \dataset{} in the public policy domain.
There are some existing domain specific datasets, including PubMedQA and BioAsq \cite{bioasq, jin2019pubmedqa} in medical domain, UDC \cite{lowe2016ubuntu} in computer software domain, QASPER \cite{dasigi2021dataset} in academic paper domain, PrivacyQA and PolicyQA \cite{ahmad-etal-2020-policyqa, ravichander-etal-2019-question} in legal domain, etc. PrivacyQA and PolicyQA have similar context as \dataset{}, but most of the questions do not require compositional reasoning and the answers are short text spans. We use a corpus in the public policy domain because it is easy to understand by non-experts while being complex enough to support challenging questions. \dataset{} is not designed to be a domain specific dataset. 

\section{The Task}
In our task, the model is provided with a long document that describes a public policy, a question about this document, and a user scenario. The model is asked to read the document and find all answers and their conditions if any.

\subsection{Corpus}
Documents in \dataset{} describe public policies in the UK, e.g. ``Apply for Visitor Visa''\footnote{\url{https://www.gov.uk/standard-visitor-visa}} or ``Punishment of Driving Violations''. Each document covers a unique topic and the contents are grouped into sections and subsections. Contents in the same section are closely related but may also be referred in other sections. We create \dataset{} in this domain because these documents are rather complex with internal logic, 
yet annotators are familiar with the content so they can ask natural yet challenging questions, compared to formal legal or financial documents with more sophisticated terms and language.

\subsection{Input} 
The input to a reading comprehension model consists of a document, a question, and a user scenario:
\begin{itemize}
    \item 
    A \textit{document} describes a public policy in the UK. 
    Content of a document is coherent and hierarchical, structured into sections and subsections. Documents are crawled from the website and processed by serializing the DOM trees of the web pages into lists of HTML elements with tags, such as <h1>, <p>, <li>, and <tr>. Please see more information in \S \ref{sec:data-documents}.
    \item A \textit{question} asks about a specific aspect of the document, such as eligibility or other aspects with ``how'', ``when'', ``what'', ``who'', ``where'', etc. Questions are relevant to the content in the document, even though they may be ``not answerable''.
    \item A \textit{user scenario} provides background information for the question. Some information will be used to restrict the answers that can be possibly correct. Not all information in the user scenario is relevant because they are written by crowd source workers without seeing the full document or knowing the answers. Information in the scenario is also likely to be incomplete. This setup simulates the real information seeking process of having both irrelevant and incomplete information.
\end{itemize}

\subsection{Output} 
A reading comprehension model is asked to predict the answers and the list of conditions if there is any.
\begin{itemize}
    \item An \textit{answer} to the question has three different types: (1) ``yes'' or ``no'' for questions such as ``Can I get this benefit?''; (2) an extracted text span for questions asking ``how'', ``when'', ``what'', etc.; (3) ``not answerable'' if an answer does not exist in the document. Since the information to get a definite answer is sometimes incomplete, besides predicting the answers, the model is asked to identify their conditions. 
    \item A \textit{condition} contains information that must be satisfied in order to make the answer correct but is not mentioned in the user scenario. In \dataset{}, we restrict a condition to be one of the HTML elements in the document instead of the exact extracted text span.\footnote{We argue that selecting HTML elements as conditions is already very challenging (see experimental results in \S \ref{sec:results}) and leave extracting the exact text spans as future work.}
     Selected conditions are then evaluated as a retrieval task with F1 at the element level, i.e. the model should retrieve all HTML elements with unsatisfied information to get a perfect F1 score. If no condition is required, the model must return an empty list. Please see \S \ref{sec:eval-metric} for more details on evaluation.
\end{itemize}

\subsection{Evaluation} \label{sec:eval-metric}
We evaluate performance of models on the \dataset{} dataset as a reading comprehension (RC) task. Answers are measured with exact match (EM) and F1. Some questions have multiple answers. The model should correctly predict all possible answers to get the full score. Since the order of answers does not matter, to compute the metrics, we compare all possible permutations of the predicted answers to the list of correct answers. We take the best result among all permutations as the result for this example. Let $\{\hat{a}_1, \dots, \hat{a}_m\}$ be the list of predicted answer and $\{a_1, \dots, a_n\}$ the reference answers.
The EM of the predicted answers is
\begin{align}\label{eq:metric}
    \textnormal{EM} = \max_{\{\tilde{a}_1, \dots, \tilde{a}_m\}} ~\frac{1}{n} \sum_{i=1}^{\min(m, n)} s_{em}(\tilde{a}_i, a_i) \cdot \gamma_{m,n}
\end{align}
$$
\gamma_{m,n} = \left\{ 
  \begin{array}{ l l }
    e^{1 - m/n} & \quad \textrm{if } m > n \\
    1                 & \quad \textrm{if } m \leq n
  \end{array}
\right.
$$

\smallskip
\noindent where $\{\tilde{a}_1, \dots, \tilde{a}_m\}$ is a permutation of the predicted answers $\{\hat{a}_1, \dots, \hat{a}_m\}$ and $s_{em}(\cdot, \cdot)$ is the scoring function that measures EM between two text spans. $\gamma_{m,n}$ is a penalty term that is smaller than $1$ if more answers than the reference answers are predicted, i.e. $m > n$. We compute token-level F1 in the similar way using the scoring function $s_{f1}(\cdot, \cdot)$ on the extracted answer spans. For not answerable questions, EM and F1 are 1.0 if and only if no answer is predicted.

We additionally measure the performance of answers with conditions. We adopt the same permutation strategy as above, except that the scoring function will also take into account the accuracy of predicted conditions. Let $\hat{C}_i$ be the set of predicted conditions for the predicted answer $\hat{a}_i$ and $C_i$ be the oracle conditions for the answer $a_i$. The new scoring function for the predicted answer with conditions is
$$
s_{em + c}(\tilde{a}_i, \tilde{C}_i, a_i, C_i) = s_{em}(\tilde{a}_i, a_i) \cdot \textnormal{F1} (\hat{C}_i, C_i)
$$
\noindent where $\textnormal{F1}(\cdot, \cdot)$ measures the accuracy of the set of predicted conditions at HTML element level. Recall that conditions are restricted to select from HTML elements in the document. $\textnormal{F1} (\hat{C}_i, C_i)$ equals to 1 if and only if all required conditions are selected. This is different from $s_{f1}(\cdot, \cdot)$ that measures token level F1 of the extracted answers.  
If the answer does not require any conditions, the model should predict an empty set. We simply replace the scoring function $s_{em}(\cdot, \cdot)$ in Eq. \ref{eq:metric} with $s_{em+c}(\cdot, \cdot)$ to compute EM with conditions.

\section{Data Collection}
\subsection{Documents}\label{sec:data-documents}
Documents are originally presented on the UK government website in the HTML format. We crawled the pages from the website and processed it to only keep the crucial tags, that include: 
\begin{itemize}
    \item Headings <h1, h2, h3, h4>: We keep headings at different levels. This can be used to identify the hierarchical structure in the documents.
    \item Text <p>: This tag is used for general contents. We replace descriptive tags, e.g. <strong>, with the plain tag <p> for simplicity.
    \item List <li>: We keep the tags for list items, but drop their parent tags <ul> or <ol>. We observe that very few ordered lists (<ol>) have been used in the dataset, so we will not distinguish them. 
    \item Table <tr>: Again, we drop their parent tags <table> to simplify the document format. We further remove the <td> and <th> tags from cells and concatenate cells in the same row with the separation of `` | ''. 
\end{itemize}

\noindent A processed document contains a list of strings that starts with a tag, follows with its content, and ends with the tag, e.g. [``<h1> Overview </h1>'', ``<p> You can apply for ... </p>'', $\dots$]. 

We drop some common sections that do not contain any crucial information, e.g. ``How to Apply'', to make sure that questions are specific to the topic of the documents. We further require that the document should contain at least 3 sections. We end up with 652 documents as our corpus. 
The average length of the documents 1358 tokens with 
a max length of 9230 tokens.

\subsection{Questions}
We collect questions from crowd source workers on Amazon Mechanical Turk. To encourage workers asking questions not be restricted to a specific piece of text, we hide the full document but instead provide a snippet of the document to the workers. A snippet includes a table of content that contains section and subsection titles (from <h1> and <h2> tags), and the very first subsection in the document that usually provides a high level overview of the topic. The snippet helps workers to familiarize themselves with the topic of this document so they can ask closely relevant questions. We observe that restricting the geographic location of workers to the UK can significantly improve the quality of questions because local residents are more familiar with their policies.

We ask the workers to perform three sub-tasks when coming up with the questions. First, we ask the workers to provide three attributes that can identify the group of people who may benefit from or be regulated by the policy discussed in the document. Second, they are asked to come up with a scenario when they will want to read this document and a question about what they would like to know. Third, workers are asked to mark which attributes have been mentioned in their question and scenario. When assessing the annotation quality, we find that asking workers to provide attributes makes the questions and scenarios much more specific, significantly improving the quality of the dataset. 

We assign 3 workers to documents with four or more sections and 2 workers to documents with three sections. Each worker is asked to give two questions and the two questions have to be diverse. We collect 3617 questions in this stage.

\subsection{Find Answers}
We hire another group of workers to work on the answer portion of this task. Finding answers is very challenging to crowd source workers because it requires the workers to read the full document carefully to understand every piece of information in the document. We provide one-on-one training for the workers to teach them how to select supporting evidences, answers, and conditions.

Workers are asked to perform three sub-tasks. The first step is to select supporting evidences from the document. Supporting evidences are HTML elements that are closely related to the questions, including elements that have content that directly justify the answers and the ones that will be selected as conditions in the next step. In the second step, workers are asked to type answers and select associated conditions. Workers can input as many answers as possible or mark the question as ``not answerable''. For each answer, they can select one or more supporting evidences as the answer's conditions if needed. Workers are asked not to select conditions if there is sufficient information in the scenario to answer the question. We give workers permission to slightly modify the questions or scenarios if the questions are not clearly stated, or they can mark it as a bad question (different from not answerable) so we will drop it from the dataset.


We additionally perform a revise step to improve the annotation quality. We provide the union of selected evidences and answers from multiple annotations of a question to an additional group of annotators and let them deselect unrelated evidences and merge answers. As the amount of information provided to workers at this step is significantly less than in the previous answer selection stage, the annotation quality improves significantly. We end up with 3102 questions with annotated answers.

\begin{table*}[]
\small
\centering
\begin{tabular}{llll}
\toprule
\textbf{Type}                                                                                            & \textbf{Scenario \& Question}                                                                                                                                                                                                                                           & \multicolumn{2}{l}{\textbf{Answer w/ {[}\textit{Conditions}{]}}}                                                                                                                                                                                                                                          \\ \midrule
Single answer                                                                                    & \begin{tabular}[c]{@{}l@{}}\textbf{Scenario}: "My father has recently appealed for\\ a traffic ticket."\\ \textbf{Question}: "How long will it take to get a decision?"\end{tabular}                                                                                 & \multicolumn{2}{l}{$\bigcdot$ "4 weeks"}                                                                                                                                                                                                                                       \\ \midrule
\begin{tabular}[c]{@{}l@{}}Single answer \\ w/ conditions\end{tabular}                           & \begin{tabular}[c]{@{}l@{}}\textbf{Scenario}: "I applied to cut down a tree on \\ my land but it was rejected 20 days ago"\\ \textbf{Question}: "Am I still able to appeal against it?"\end{tabular}                                                                 & \multicolumn{2}{l}{\begin{tabular}[c]{@{}l@{}}$\bigcdot$ "yes"\\ ~~{[}\textit{"\textless{}p\textgreater{}You can appeal before the date the tree} \\ ~~~~\textit{replacement notice comes into effect.\textless{}/p\textgreater{}"}{]}\end{tabular}}                                                   \\ \midrule
\multirow{3}{*}{Multiple answers}                                                                & \multirow{3}{*}{\begin{tabular}[c]{@{}l@{}}\textbf{Scenario}: "I will get my first paycheck tomorrow."\\ \textbf{Question}: "What information should be on \\ my pay split?"\end{tabular}}                                                                           & \multicolumn{2}{l}{$\bigcdot$ "earnings before and after any deductions"}                                                                                                                                                                                                      \\
                                                                                                 &                                                                                                                                                                                                                                                    & \multicolumn{2}{l}{\begin{tabular}[c]{@{}l@{}}$\bigcdot$ "the amount of any deductions"\end{tabular}}                                                                                                                                 \\
                                                                                                 &                                                                                                                                                                                                                                                    & \multicolumn{2}{l}{$\bigcdot$ "the number of hours you worked"}                                                                                                                                                                                                                \\ \midrule
\multirow{2}{*}{\begin{tabular}[c]{@{}l@{}}Multiple yes/no \\ w/ conditions\end{tabular}}        & \multirow{2}{*}{\begin{tabular}[c]{@{}l@{}}\textbf{Scenario}: "I am looking at buying a new build \\ home. I am 26 and a first-time buyer."\\ \textbf{Question}: "Am I eligible to get an Equity Loan?"\end{tabular}}                                                & \multicolumn{2}{l}{\begin{tabular}[c]{@{}l@{}}$\bigcdot$ "yes"\\ ~~{[}\textit{"\textless{}li\textgreater{}able to afford the fees and interest\textless{}li\textgreater{}"}, \\ ~~~~\textit{"\textless{}li\textgreater{}sold by an eligible homebuilder\textless{}/li\textgreater{}"}{]}\end{tabular}}         \\
                                                                                                 &                                                                                                                                                                                                                                                    & \multicolumn{2}{l}{\begin{tabular}[c]{@{}l@{}}$\bigcdot$ "no"\\ ~~{[}\textit{"\textless{}p\textgreater{}You can not apply if you had any} \\ ~~~~\textit{form of sharia mortgage finance\textless{}/p\textgreater{}"}{]}\end{tabular}}                                                                       \\ \midrule
\multirow{2}{*}{\begin{tabular}[c]{@{}l@{}}Multiple extractive \\ w/ conditions\end{tabular}}    & \multirow{2}{*}{\begin{tabular}[c]{@{}l@{}}\textbf{Scenario}: "I always walk my labrador in open \\ spaces. I forgot to clean up his mess yesterday."\\ \textbf{Question}: "How much can I be fined for this?"\end{tabular}}                                         & \multicolumn{2}{l}{\begin{tabular}[c]{@{}l@{}}$\bigcdot$ "\$100"\\ ~~{[}\textit{"\textless{}li\textgreater{}\$100 on the spot\textless{}/li\textgreater{}"}{]}\end{tabular}}                                                                                                                    \\
                                                                                                 &                                                                                                                                                                                                                                                    & \multicolumn{2}{l}{\begin{tabular}[c]{@{}l@{}}$\bigcdot$ "up to \$1000"\\ ~~{[} \textit{"\textless{}li\textgreater{}up to \$1,000 if it goes to court\textless{}/li\textgreater{}"}{]}\end{tabular}}                                                                                             \\ \midrule
\multirow{4}{*}{\begin{tabular}[c]{@{}l@{}}Multiple extractive \\ one w/ condition\end{tabular}} & \multirow{4}{*}{\begin{tabular}[c]{@{}l@{}}\textbf{Scenario}: "I am about to apply for a Parent of a \\ Child Student Visa to stay with my child for \\ a year in the UK" \\ \textbf{Question}: "What documents are needed to apply \\ for this visa?"\end{tabular}} & \multicolumn{2}{l}{$\bigcdot$ "a current passport or other travel document"}                                                                                                                                                                                                   \\
                                                                                                 &                                                                                                                                                                                                                                                    & \multicolumn{2}{l}{$\bigcdot$ "proof that you have enough fund"}                                                                                                                                                                                                               \\
                                                                                                 &                                                                                                                                                                                                                                                    & \multicolumn{2}{l}{$\bigcdot$ "proof of permanent address outside the uk"}                                                                                                                                                                                                     \\
                                                                                                 &                                                                                                                                                                                                                                                    & \multicolumn{2}{l}{\begin{tabular}[c]{@{}l@{}}$\bigcdot$ "your tuberculosis (tb) test results"\\ ~~{[}\textit{"\textless{}li\textgreater{}your tuberculosis (TB) test results} \\ ~~~~\textit{if you are from a country where you} \\ ~~~~\textit{have to take the TB test\textless{}/li\textgreater{}"}{]}\end{tabular}} \\ \bottomrule
\end{tabular}
\caption{\small Example of questions in \dataset{}. Text pieces that follows the answers in the brackets are [\textit{conditions}]. Some answers are deterministically correct so they are not followed by conditions.}
\label{tab:examples}
\end{table*}

\subsection{Move Conditions to Scenario}
To encourage the model of learning subtle difference in user scenarios that affects the answers and conditions, we create new questions by modifying existing questions with conditional answers by moving one of the conditions to their scenarios.


Specifically, we show the workers the original questions, scenarios, and the annotated answers and conditions. Evidences are also provided for workers to get them familiar with the background of the questions and reasoning performed to get the original answers. Workers are asked to pick one of the conditions and modify the original scenario to reflect this condition. The modified questions and scenarios are sent back to the answering stage to get their annotations. We randomly select a small portion of the questions that have conditional answers as inputs to this stage so as to not affect the original distribution of the dataset. We collected 325 additional examples from this stage.

\subsection{Train / Dev / Test Splits}
We partition the dataset by documents to prevent leaking information between questions from the same document. The dataset contains 436 documents and 2338 questions in the training set, 59 documents and 285 questions in the development set, and 136 documents and 804 questions in the test set.

\section{Dataset Analysis}
The dataset consists of yes/no questions and extractive questions. Questions may contain one or more answers, with or without conditions. The statistics of the questions are shown in Table \ref{tab:stats}.

\begin{table}[]
\small
\centering
\begin{tabular}{llc}
\toprule
 & \textbf{Type} & \textbf{\#} \\
\midrule
\multirow{2}{*}{Answer type}       & yes / no         & 1751 \\
                                  & extractive    & 1527 \\
\midrule
\multirow{2}{*}{Condition type}       & deterministic & 2475 \\
                                  & conditional   & 803  \\
\midrule
\multirow{2}{*}{Number of answers} & single        & 2526 \\
                                  & multiple      & 752  \\
\midrule
--                     & not answerable             & 149  \\
\bottomrule
\end{tabular}
\caption{\small Statistics on different types of questions.}
\label{tab:stats}
\end{table}

\noindent\textbf{Answer type} Among all the answerable questions, 1751 questions have yes/no answers while the other 1527 questions have extractive answers. 1161 of the yes/no questions have the answer ``yes'', 712 questions have answer ``no'', and 122 questions have both answers ``yes'' and ``no'' under different conditions. Please see the example in Table \ref{tab:examples}. The average length of the extract answers is 6.36 tokens.

\noindent\textbf{Condition type} 803 questions have conditional answers. 390 out of the 803 questions have one answer, but this answer is only correct if the conditions are satisfied. 173 questions have multiple answers, each have their own conditions, i.e. the answers are different if different conditions apply. The rest 240 questions also have multiple answers, but some of the answers require conditions while other don't. See examples in Table \ref{tab:examples}. A total of 1090 answers from 803 questions have conditions, among which 672 answers have only one condition and 418 answers have multiple conditions.

\noindent\textbf{Number of answers} Besides questions that have different answers under different conditions, 339 questions have multiple deterministic answers.

\begin{table*}[]
\small
\centering
\begin{tabular}{lcccccccc}
\toprule
          & \multicolumn{2}{c}{Yes / No} & \multicolumn{2}{c}{Extractive} & \multicolumn{2}{c}{Conditional} & \multicolumn{2}{c}{Overall} \\
          & answers      & w/ conditions      & answers      & w/ conditions   & answers      & w/ conditions*  & answers      & w/ conditions \\ \midrule
majority  & 62.2 / 62.2  & 42.8 / 42.8  & -- / --  & -- / --   & -- / --  & -- / --  & -- / --  & -- / -- \\
ETC-pipeline   & 63.1 / 63.1  & 47.5 / 47.5   & 8.9 / 17.3     & 6.9 / 14.6 & 39.4 / 41.8  & 2.5 / 3.4  & 35.6 / 39.8   & 26.9 / 30.8     \\
DocHopper & \textbf{64.9 / 64.9}  & \textbf{49.1 / 49.1}   & 17.8 / 26.7    & 15.5 / 23.6  & 42.0 / 46.4  & 3.1 / 3.8   & 40.6 / 45.2   & 31.9 / 36.0     \\
FiD       & 64.2 / 64.2  & 48.0 / 48.0   & \textbf{25.2 / 37.8}    & \textbf{22.5 / 33.4}  & \textbf{45.2 / 49.7}  & \textbf{4.7 / 5.8}    & \textbf{44.4 / 50.8} & \textbf{35.0 / 40.6}  \\ \midrule
human  & 91.4 / 91.4  & 82.3 / 82.3 &  72.6 / 84.9 & 62.8 / 69.1  & 74.7 / 86.9  & 48.3 / 56.6   & 82.6 / 88.4 & 73.3 / 76.2 \\
\bottomrule
\end{tabular}
\caption{\small Experiment results on \dataset{} (EM / F1). Numbers are obtained by re-running the open-sourced codes of the baselines. ``majority'' reflects the accuracy of always predicting ``yes'' without conditions. *See discussion in text.}
\label{tab:exp-results}
\end{table*}

\section{Evaluation}

\subsection{Baselines}
Evaluating existing models on \dataset{} is challenging. In addition to predicting  answers to questions, the \dataset{} task also asks the model to find the answers' conditions if any of them applies. To the best of our knowledge, no existing model fits the purpose of this task. We modified three competitive QA models as baselines to the \dataset{} dataset. In addition to the new form of answers, traditional reading comprehension models also face the challenge that the context of questions in \dataset{} is too long to fit into the memory of many Transformer-based models like BERT and RoBERTa \cite{devlin2019bert,liu2019roberta}.  The baseline models we implemented are described below.

\noindent \textbf{ETC-pipeline}: ETC \cite{ainslie2020etc} is a pretrained Transformer-based language model that is designed for longer inputs (up to 4096 tokens). ETC achieved the state-of-the-art on several challenging tasks, e.g. HotpotQA and WikiHop \cite{yang2018hotpotqa, welbl2018constructing}. Since vanilla ETC cannot jointly predict answers and conditions, we designed a two stage pipeline to run ETC on \dataset{}.

In the first stage, ETC is trained as a normal reading comprehension model to predict answers from the document by jointly encoding the concatenated tokens from questions and documents. 
Since ETC cannot fit the entire document (up to 9230 tokens) into its memory, we adopt a sequential reading approach that reads one section at a time. The answer with the highest probability among all sections will be considered as the final answer. We append three special tokens ``\texttt{yes}'', ``\texttt{no}'', and ``\texttt{not answerable}'' for the yes/no and not answerable questions. Since it is not clear how to extract multiple answers with the Transformer-based extractive QA model, we restrict to the number of predicted answers to one. 

The second stage in the pipeline is to select conditions. Questions, answers, and documents are concatenated together into a single input and encoded by ETC. We then use the embeddings of global tokens in ETC to predict conditions. Since the number of conditions for an answer is unknown, we train ETC to label its global tokens (each represent a candidate condition) as a binary classification target. The threshold of selecting conditions is tuned as a hyper-parameter.

\noindent \textbf{DocHopper}: DocHopper \cite{sun2021endtoend} is an iterative attention method for reading long documents to answer multi-hop questions. The model iteratively attends to information at different levels in the document to gather evidences to predict the final answers. We modify the iterative process in DocHopper for the purpose of this task: specifically, DocHopper is trained to run three iterative attention steps: (1) attend to the supporting evidences; (2) attend to the sentence that contains the answer; and (3) attend to the conditions. Since the query vector in each attention step is updated with information from the previous steps, conditions attended at the third step are aware of the previously predicted answers. Similar to the ETC pipeline, we restrict the model to predict one answer for each question. The condition selection step in DocHopper is also trained with binary classification loss. Different from the ETC pipeline, the three attention steps are jointly optimized.

\noindent \textbf{FiD}: FiD \cite{izacard2021leveraging} is a generative model with an encoder-decoder architecture. The encoder reads multiple contexts independently and generates their embeddings. The decoder attends to all embeddings of the context to generate the final answers. In this task, we train FiD to sequentially generate the answers with conditions, i.e. $[a_1, c_{11}, c_{12}, \dots, a_2, c_{21}, c_{22}, \dots]$ where $\{a_1, \dots, a_n\}$ are the correct answers and $\{C_1, \dots, C_n\}$ are their conditions, i.e., $c_{ij} \in C_i$ is the $j$'th condition for the answer $a_i$. If $C_i$ is empty, the model is trained to predict ``NA'' as the only condition for the $i$'th answer.

\noindent \textbf{Human} We randomly sample 80 questions and ask human annotators to answer them. Annotators are provided with the full instructions and 10 additional annotated examples to clarify the task. We do not provide extensive training to the annotators. No additional revision step is performed on the answers.

\subsection{Results} \label{sec:results}
\begin{table*}[t]
\small
\centering

\begin{tabular}{lclll}
\toprule
                                                              \textbf{Error types}                     & \textbf{\%}      & \multicolumn{1}{l}{\textbf{Examples}}                                                                                & \multicolumn{1}{l}{\textbf{Correct answers}}                                                                             & \multicolumn{1}{l}{\textbf{Predictions}} \\ \midrule
Not answerable                                                                      & 7.6  & \begin{tabular}[c]{@{}l@{}}"Am I eligible for a tax \\ reduction?"\end{tabular}                             & not\_answerable                                                                                       & "yes"                            \\ \midrule
\begin{tabular}[c]{@{}l@{}}Wrong answer type\\ (yes/no vs. extractive)\end{tabular} & 4.2  & \begin{tabular}[c]{@{}l@{}}"How can I check if \\ this design has been \\ registered?"\end{tabular}         & \begin{tabular}[c]{@{}l@{}}"ask the intellectual\\ property office to \\ search for you"\end{tabular} & "no"                             \\ \midrule
\begin{tabular}[c]{@{}l@{}}Wrong answer\\ (yes/no)\end{tabular}                     & 19.5 & \begin{tabular}[c]{@{}l@{}}"Will it be classed as \\ a small vessel?"\end{tabular}                          & "yes"                                                                                                 & "no"                             \\ \midrule
\begin{tabular}[c]{@{}l@{}}Wrong answer\\ (extractive, right type)\end{tabular}     & 20.3 & \begin{tabular}[c]{@{}l@{}}"How many points will \\ I receive on my license?"\end{tabular}                  & "6"                                                                                                   & "3"                              \\ \midrule
\begin{tabular}[c]{@{}l@{}}Wrong answer\\ (extractive, wrong type)\end{tabular}     & 9.3  & \begin{tabular}[c]{@{}l@{}}"What is the account \\ number should I send \\ the money to?"\end{tabular}      & "12001020"                                                                                            & "hmrc"                           \\ \midrule
\begin{tabular}[c]{@{}l@{}}Correct answer\\ w/ wrong conditions\end{tabular}                                                                     & 14.4 & \begin{tabular}[c]{@{}l@{}}"Can I still send simpler \\ annual accounts as a \\ micro-entity?"\end{tabular} & \begin{tabular}[c]{@{}l@{}}"yes", \\ {[}"\textit{\$316,000 or less} \\ \textit{on its balance sheet"}{]}\end{tabular}  & "yes", {[}{]}                             \\ \midrule
Partial answer                                                                      & 24.5 & \begin{tabular}[c]{@{}l@{}}"What will not need to \\ be repeated for each trip?"\end{tabular}               & \begin{tabular}[c]{@{}l@{}}"a microchip", \\ "rabies vaccination"\end{tabular}                        & "a microchip"                  \\ \bottomrule
\end{tabular}
\caption{\small Error analysis on the predictions of the best performed model (FiD). The percentage is the fraction of errors made in that category over all errors.}
\label{tab:error}
\end{table*}

Experiment results are shown in Table \ref{tab:exp-results}. We report the numbers on yes/no questions and extractive questions separately. The numbers in Table \ref{tab:exp-results} show that the \dataset{} task is very challenging---the performance of the best model on yes/no questions is 64.9\% (marginally higher than always predicting the majority answer ``yes''), and the performance on extractive questions is 25.2\% EM. FiD has the best performance on extractive questions because FiD can predict multiple answers while ETC-pipeline and DocHopper only predict one. 

The performance drops significantly if answers and conditions are jointly evaluated. The best performance on jointly evaluating answers and conditions (``w/ conditions'') in Table \ref{tab:exp-results} is only 49.1\% EM for yes/no questions and 22.5\% EM for extractive questions. Even worse, this best performance is obtained when \textit{no} condition is selected, i.e. the threshold of selecting conditions is set to $1.0$. 
The difficulty of selecting conditions is more obvious if we focus on the subset of questions that have at least one conditional answer. The accuracy drops by more than 90\% if answers and conditions are jointly evaluated.\footnote{The EM/F1 w/ conditions* is non-zero on this subset of questions even if no condition is ever selected, because some questions have both conditional and deterministic answers.} 

\begin{figure}[]
    \centering
    \includegraphics[width=0.35\textwidth]{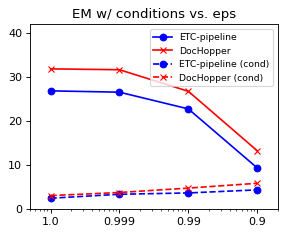}
    \caption{\small EM of answers with conditions with different thresholds of confidence (eps) on conditions. Dotted lines represent experiment results on the subset of questions that have conditional answers.}
    \label{fig:em-vs-eps}
\end{figure}

We also study how the threshold on the confidence scores of selecting conditions affects the evaluation results. Results are shown in Figure \ref{fig:em-vs-eps}. As we decrease the threshold for selecting conditions, the EM with conditions on the subset of questions that have conditional answers slightly improves, but the overall EM with conditions drops dramatically due to the false positive conditions.
FiD is a generative model so we can not evaluate it in the same way. In our evaluation, predictions from the best performing FiD checkpoint also do not select any conditions. 

Table \ref{tab:exp-results-cond} shows the best results on the subset of questions that have conditional answers. Hyper-parameters are tuned on the subset of questions. We could possibly get better results on questions with conditional answers with threshold $\epsilon < 1.0$, but the improvement is still marginal.

\begin{table}[]
\small
\centering
\begin{tabular}{lcc}
\toprule
          & Best Overall      & Best Conditional      \\ \midrule
ETC-pipeline     & 2.5 / 3.4    & 4.4 / 4.6     \\
DocHopper   & 3.1 / 3.8  & \textbf{5.9 / 7.1}   \\
FiD        & \textbf{4.7 / 5.8}   & 4.7 / 5.8    \\
\bottomrule
\end{tabular}
\caption{\small EM/F1 w/ conditions on the subset of questions in \dataset{} with \textit{conditional answers}. ``Best Overall'' uses the best checkpoints/hyper-parameters on the full dataset, while ``Best Conditional'' uses the best ones on the subset of questions.}
\label{tab:exp-results-cond}
\end{table}

\subsection{Error Analysis}

We manually check 200 examples in the prediction of the best performed model FiD and label the type of errors made. The numbers are shown in Table \ref{tab:error}. The most errors are made when only a subset of correct answers is predicted. This is due to the fact that the model (FiD) has a tendency to predict one answer for each question. The second most common errors are made by predicting answers with the correct type but wrong value. Such errors are commonly made by reading comprehension models in many tasks. The model made a lot of errors in yes/no questions because they consist of around 50\% of the questions. The model is good at distinguishing yes/no questions and extractive question as producing the wrong kind of answer only makes up of 4.2\% of the errors.


\section{Conclusion}

We propose a challenging dataset \dataset{} that contains questions with conditional answers. The dataset requires models to understand complex logic in a document in order to find correct answers and conditions to the questions. Experiments on state-of-the-art QA models show that their overall performance on \dataset{} is relatively poor. This also suggests that current QA models lack the reasoning ability of identifying answers with conditions. We hope that this dataset will stimulate further research in building QA models for answering questions with conditions and building broader NLP models with better reasoning abilities.

\section{Ethics Statements}
This dataset should be ONLY used for NLP research purpose. Questions are artificial and do not contain any personal information. Answers are NOT provided by legal professionals and should NOT be used for any legal purposes.

\section{Acknowledgement}
This work was supported in part by the NSF IIS1763562,  ONR Grant N000141812861, Google Research.
We would also like to thank Vijay A. Saraswat \texttt{<Vijay.Saraswat@gs.com>} for valuable feedback.


\bibliographystyle{acl_natbib}




\end{document}